\pdfoutput=1

\documentclass[11pt]{article}

\usepackage{acl}

\usepackage{times}
\usepackage{latexsym}

\usepackage[T1]{fontenc}

\usepackage[utf8]{inputenc}

\usepackage{microtype}

\usepackage{inconsolata}

\usepackage{graphicx}

\usepackage{lipsum}  
\usepackage{tabularx}
\usepackage{booktabs}

\usepackage{microtype}
\usepackage{type1cm}
\usepackage[american]{babel}
\usepackage{amssymb}
\usepackage{amsmath}
\usepackage{multirow}
\usepackage{xspace}
\DeclareMathAlphabet{\mathcalbf}{OMS}{pzc}{b}{n}
\usepackage{enumitem}
\usepackage{colortbl}

\RequirePackage{color}
\definecolor{darkgray}{gray}{0.40}
\definecolor{mediumgray}{gray}{0.60}
\definecolor{lightgray}{gray}{0.95}
\definecolor{ultralightgray}{gray}{0.98}
\definecolor{forestgreen}{rgb}{0.133, 0.545, 0.133}
\definecolor{orange}{rgb}{1, 0.86, 0.74}
\definecolor{lightergreen}{rgb}{0.95, 1, 0.88}

\usepackage{graphicx}
\DeclareGraphicsExtensions{.pdf,.ai,.jpg,.png}
\setkeys{Gin}{pagebox=artbox}
\graphicspath{{./acl24-argpaca-figures/}}

\newcommand{\bsfigure}[3][]{%
    \begin{figure}[t]
        \centering
        \includegraphics[#1]{#2}
        \caption{#3}\label{#2}%
    \end{figure}
}

\newcommand{\hwfigure}[3][t!]{%
    \begin{figure*}[#1]
        \centering
        \includegraphics[scale=1.0]{#2}
        \caption{#3}\label{#2}%
    \end{figure*}
}

\RequirePackage{type1cm}
\RequirePackage{color}
\RequirePackage{soul}
\setstcolor{blue}
\definecolor{violet}{rgb}{0.5,0.0,0.5}

\newsavebox\bscombox
\newcommand{\bscom}[3][]{%
    \sbox{\bscombox}{\fontsize{8}{9}\selectfont#1#2#3}
    \noindent
    \st{#2}{\selectfont
        \color{blue}#3\ifx\\#1\\\else{\fontsize{8}{9}\selectfont\color{violet}[#1]}\fi
    }
}

\usepackage{tikz}

\begin{document}

\title{ArgInstruct: Specialized Instruction Fine-Tuning \\ for Computational Argumentation}




\author{
	Maja Stahl\Thanks{Equal contribution.} \\
	Leibniz University Hannover \\
	\texttt{m.stahl@ai.uni-hannover.de} \And
	Timon Ziegenbein\footnotemark[1] \\
	Leibniz University Hannover \\
	\texttt{t.ziegenbein@ai.uni-hannover.de} \AND
	Joonsuk Park\Thanks{Equal advising.} \\
	University of Richmond \\
	\texttt{park@joonsuk.org} \And
	Henning Wachsmuth\footnotemark[2] \\
	Leibniz University Hannover \\
	\texttt{h.wachsmuth@ai.uni-hannover.de}
}




\date{}

\maketitle

\begin{abstract}

Training large language models (LLMs) to follow instructions has significantly enhanced their ability to tackle unseen tasks. However, despite their strong generalization capabilities, instruction-following LLMs encounter difficulties when dealing with tasks that require domain knowledge. This work introduces a \emph{specialized instruction fine-tuning} for the domain of computational argumentation (CA). The goal is to enable an LLM to effectively tackle any unseen CA tasks while preserving its generalization capabilities. Reviewing existing CA research, we crafted natural language instructions for 105 CA tasks to this end. On this basis, we developed a CA-specific benchmark for LLMs that allows for a comprehensive evaluation of LLMs' capabilities in solving various CA tasks. We synthesized 52k CA-related instructions, adapting the self-instruct process to train a CA-specialized instruction-following LLM. Our experiments suggest that CA-specialized instruction fine-tuning significantly enhances the LLM on both seen and unseen CA tasks. At the same time, performance on the general NLP tasks of the SuperNI benchmark remains stable.

\end{abstract}

\section{Introduction}

\bsfigure{scheme}{Comparison of fine-tuning methods: (a) Optimizing an LLM for a CA task on input-output pairs. (b) Making an LLM instruction-following on highly diverse tasks. (c) Our method: Making an LLM an instruction-following CA specialist on diverse CA-specific tasks.}

Large language models (LLMs) have proven effective for various NLP tasks, including several tasks from computational argumentation (CA), the computational analysis and synthesis of natural language arguments \cite{chen-etal-2024-exploring-potential}. Initially, it was common to fine-tune pretrained LLMs on input-output pairs for a task \cite{devlin:2019,radford:2019}. Figure~\ref{scheme}(a) illustrates such \emph{task-specific fine-tuning} for the mining of claims and premises from student essays. In contrast, recent LLMs are often \emph{instruction fine-tuned} by exposing them to highly diverse tasks%
\footnote{Here, the term \emph{task} refers to a natural language instruction along with one or more input-output pairs that provide contextual guidance \cite{mishra-etal-2022-cross, wang-etal-2022-super}.} 
\cite{ouyand-etal-2022-training-large,taori-etal-2023-stanford-alpaca}, as shown in Figure~\ref{scheme}(b).
This enables them to generate responses aligned with specific task requirements described in the instruction \cite{wang-etal-2022-super, wang-etal-2023-self-instruct}. 

However, despite their strong generalization abilities, instruction-following LLMs often struggle to solve tasks that require domain knowledge \cite{lecler-etal-2023-revolutionizing, nascimento-etal-2023-do, yang-etal-2023-empower}. This limitation results from the principle of general instruction fine-tuning to prioritize generalizability over specialization. It affects CA tasks in particular, as they often center around sophisticated context-related concepts from argumentation theory \cite{wachsmuth-etal-2024-argument}.

Specifically, CA research in NLP focuses on the mining, assessment, and generation of natural language arguments \cite{stede:2018}. Despite recent advancements in LLMs, tackling CA tasks remains challenging \cite{chen-etal-2024-exploring-potential} due to their context-dependent 
specificities (e.g., in newspaper articles vs. social media posts) \cite{habernal-etal-2014-argumentation} and their subjectivity (e.g., in assessing argument quality) \cite{wachsmuth-etal-2017-computational, romberg-2022-perspective}. In fact, providing context- and argumentation-specific knowledge, such as details about the debate setting and definitions of argumentative concepts, has been 
stressed to be important for task performance \cite{lauscher-etal-2022-scientia}.

In this paper, we study the impact of conflating the two learning paradigms of task-specific fine-tuning and general instruction fine-tuning. That is, we introduce the idea of \emph{specialized instruction fine-tuning} by combining a highly diverse set of general tasks with a diverse set of tasks specific to a given task domain, CA in our case. The goal is to obtain an LLM that is highly proficient in CA while being agnostic to the particular CA task it encounters and maintaining generalization capabilities. 

We hypothesize that providing an LLM with argumentation-specific knowledge during instruction fine-tuning in a way that enables joint learning of representations across tasks is key to addressing the limitations of both general-purpose and task-specific LLMs in this domain. Through \emph{CA-specialized instruction fine-tuning} (Figure~\ref{scheme}(c)), we enhance the LLM's ability to mine argument structure, assess argument quality, and generate arguments across CA contexts, ensuring both accuracy and versatility for the full spectrum of argumentation. We expect this to require many and diverse CA tasks, 
exceeding what can be achieved by merely combining existing tasks.

Towards the outlined goal, we create a large CA-specific instruction fine-tuning dataset. Starting from 105 seed tasks, derived from a total of 30 
argumentation corpora, we follow the self-instruct process of \citet{wang-etal-2023-self-instruct} to automatically generate a diverse set of 52k CA-specific tasks (instructions plus input-output pairs). By combining these tasks with general instruction fine-tuning data in various ways, we train instruction-following Gemma \cite{gemma-team-etal-2024-gemma} variants for CA.

The 
seed tasks serve as a new CA benchmark. Our experiments suggest that our specialized 
instruction fine-tuning method (dubbed \emph{ArgInstruct}) successfully generalizes toward unseen CA tasks, outperforming a wide range of competitive instruction-following LLMs in a zero-shot setting. Moreover, we demonstrate on SuperNI \cite{wang-etal-2022-super} 
that the LLM's general instruction-following abilities remain despite specialization.

Altogether, this paper's main contributions are:
\begin{itemize}
    \setlength\itemsep{-0pt}
    \item A general method for specialized instruction fine-tuning, instantiated for CA
    \item An extensive dataset for CA-specific instruction fine-tuning and benchmarking of LLMs
    \item Empirical evidence that our CA-specialized instruction-finetuning effectively enhances an LLM's generalizability for unseen CA tasks%
    \footnote{Our dataset and experiment code can be found under: \url{https://github.com/webis-de/ACL-25}}
\end{itemize}

\section{Related Work}
The computational analysis and synthesis of arguments in natural language, often referred to as computational argumentation (CA), has its roots in a long history of philosophical research \cite{aristotle:2007}, which has gained significant attention from the NLP community in recent years. The three main CA research areas frequently covered are argument mining \cite{park-cardie:2014:W14-21,boltuzic-snajder-2014-back,stab-gurevych-2017-parsing}, argument assessment \cite{persing-ng-2015-modeling,wachsmuth-etal-2017-computational,gretz-etal-2020-large}, and argument generation \cite{syed-etal-2021-generating,schiller-etal-2021-aspect,wachsmuth-etal-2018-argumentation}. 

\hwfigure{approach}{Overview of our methodology: We manually craft CA-specific seed tasks and prompt an LLM to generate new CA-specific tasks in a loop by (1) generating new instructions, (2) filtering them for CA relevance and novelty, and (3) generating corresponding instances. (4) After postprocessing, the generated CA-specific tasks from the task pool are combined with existing general tasks to specialize an LLM for CA using instruction fine-tuning.}

Although the contributions to each area are plentiful, most works focus on one or a few tasks within or across the areas. Many methods rely on supervised learning and single-domain datasets, limiting generalizability 
\cite{waldis-etal-2024-handle}. Recently, \citet{chen-etal-2024-exploring-potential}, \citet{elaraby-etal-2024-persuasiveness}, \citet{rescala-etal-2024-language}, and \citet{cabessa-etal-2025-argument} studied the potential of LLMs to tackle a selection of mining, assessment, and generation tasks. They obtained promising results, showing that LLMs can address multiple CA tasks, sometimes even without explicit training.
Beyond that, \citet{wachsmuth-etal-2024-argument} propose to systematically instruct LLMs for argument quality assessment with argumentation-specific knowledge to enable knowledge sharing across tasks and contexts. Despite these advances, there remains a notable gap: no study has yet comprehensively evaluated LLMs across all three main CA areas or operationalized a systematic framework such as the one suggested by \citet{wachsmuth-etal-2024-argument} to tackle CA tasks holistically. 


To fill this gap, we use instruction fine-tuning as it allows training a task-agnostic LLM for CA. In general, instruction fine-tuning is just a supervised training process. The key is the data used for fine-tuning: By having instances with instructions, and by diversifying these instances as much as possible, the LLM learns that task specificities should be abstracted from while instructions should always be followed \cite{wang-etal-2024-survey}.

Over the last years, multiple instruction fine-tuning datasets have been collecting from existing NLP tasks \cite{mishra-etal-2022-cross,wang-etal-2022-super}. The datasets usually consist of natural language instructions and example instances, which are either written manually by humans \cite{sanh-etal-2021-multitask,ouyand-etal-2022-training-large,longpre-etal-2023-flan} or created synthetically \cite{honovich-etal-2023-unnatural,wang-etal-2023-self-instruct,taori-etal-2023-stanford-alpaca,chen-etal-2024-selfplay}. The importance of data diversity and selection was further emphasized by \citet{bukharin-zhao-2023-data}, \citet{li-etal-2024-quantity}, and \citet{wang-etal-2024-survey}. Besides evaluating instruction following capabilities in terms of text generation performance on unseen tasks \cite{chia-etal-2024-instructeval,dubois-etal-2024-length-controlled}, new ways of assessment were developed in which LLMs compete against each other \cite{zheng-etal-2023-judging,chiang-etal-2024-chatbot}, sometimes even replacing human annotators as evaluators \cite{zheng-etal-2023-judging}. 

Several instruction-following LLMs have been developed and evaluated on these benchmarks \cite{ouyand-etal-2022-training-large,muennighoff-etal-2023-crosslingual,chung-etal-2024-scaling}. However, standard instruction fine-tuning is purely generalization-oriented, whereas we aim to balance between generalization and CA specialization. By changing what data is used, the LLM learns that not all task-specificities should be abstracted from, but the domain specialization should be kept while following instructions. This makes the LLM a computational argumentation expert, rather than a general talk solver as in standard instruction fine-tuning.

Our method builds on the ideas of
\emph{Self-Instruct} \cite{wang-etal-2023-self-instruct} and \emph{Alpaca} \cite{taori-etal-2023-stanford-alpaca}. Both generate 52k tasks starting from a small set of manually-written seed tasks, showing that larger amounts of diverse instructions can help improve the performance of LLMs. However, we do not focus on or start from general instructions but CA-specific ones. Hence, we combine the CA-specific knowledge of the structure, quality, and writing of arguments with meta-knowledge of how to solve tasks acquired by following general instructions.

\section{Methodology}

This section presents our methodology for the training of a specialized LLM for computational argumentation (CA).
We start by deriving a seed set of instruction fine-tuning tasks from existing CA tasks and datasets. This manually-annotated seed data serves as a reliable basis for generating a large set of diverse CA-specific tasks. Combining these tasks with general NLP tasks, we then specialize an LLM using instruction fine-tuning, dubbed \emph{ArgInstruct} (\emph{Arg}umentation-specialized \emph{Instruct}ion Fine-Tuning). Whereas standard instruction fine-tuning is, by concept, fully generalization-oriented, we diversify instructions only within the task domain. The CA-specific instructions enable the LLM to deal specifically with any task from computational argumentation. By still mixing in general instruction fine-tuning data, we further achieve that generalization capabilities are widely preserved. Figure~\ref{approach} illustrates the methodology.

\subsection{Task Generation}

Following instruction fine-tuning literature \cite{mishra-etal-2022-cross,wang-etal-2022-super,wang-etal-2023-self-instruct} we define a task $T=(I,S)$ to consist of a natural language instruction $I$ and $m \geq 1$ input-output instances $S =\{x_{j},y_{j}\}^{m}_{j=1}$. To instruction fine-tune a specialized LLM for CA, we propose to create a large instruction fine-tuning dataset $\mathcal{T} = \{T_1, \ldots, T_n\}$ containing a large set of CA-specific but diverse tasks by (1) generating new instructions, (2) filtering for CA relevance and diversity, and (3) generating corresponding input-output instances.

\paragraph{Instruction Generation}

Building on research in CA, we curate a collection of seed tasks, $\mathcal{T}_0$, across the three main research areas of CA: argument mining, argument assessment, and argument generation (see Section~\ref{sec:data-datasets}). First, we manually craft a set of $k \gg 0$ natural language instructions $\mathcal{I}_{0} = \{I_1, \ldots, I_k\}$ by extracting task and term definitions from the papers and annotation guidelines. For each $I_j$, we obtain the input-output instances from the corresponding datasets to construct $\mathcal{T}_0$. 

Following self-instruct \cite{wang-etal-2023-self-instruct}, we use $\mathcal{T}_0$ as the initial CA task pool, and generate new CA-specific instructions $\mathcal{I}_i$ using a pretrained LLM. Self-instruct is an iterative process that uses prompting to create an instruction fine-tuning dataset, where an LLM generates new training data by leveraging its own previous outputs as few-shot examples. In each generation step~$i$, we randomly sample a subset $\tilde{\mathcal{I}}$ of size $l \geq 1$ from the instructions $\mathcal{I}_{0} \cup \mathcal{I}_{<i}$ in $\mathcal{T}_0$ where $\mathcal{I}_{<i}$ contains all instructions generated previously (in Section~\ref{sec:data}, we set $l = 8$). $\tilde{\mathcal{I}}$ is used as few-shot examples to generate a new set of instructions~$\mathcal{I}_i$. To ensure that the instructions in $\mathcal{I}_{<i}$ remain focused on CA, we modify the self-instruct prompt to: ``Come up with a series of \emph{computational argumentation} tasks.''

\paragraph{Instruction Filtering}

To filter the generated instructions $\mathcal{I}_{i}$ by CA relevance, we utilize the same LLM using the prompt ``Does the following task fall into the field of computational argumentation?'' together with few-shot examples. For this, we randomly sample CA-specific examples $\mathcal{I}_{+}$ from our seed instructions $\mathcal{I}_{0}$ as positive examples and not CA-specific instructions $\mathcal{I}_{-}$ from established instruction following datasets as negative examples and append them to the prompt. Following the instruction filtering used in self-instruct \cite{wang-etal-2023-self-instruct}, a generated instruction that has been deemed CA-relevant is added to the CA task pool only when its ROUGE-L F$_1$-score similarity with any existing instruction from $\mathcal{I}_{0} \cup \mathcal{I}_{<i}$ falls below a predefined threshold $\tau$ (we use $\tau = 0.7$ below). The instruction generation and filtering process is repeated until we reach a predefined number of generated instructions. In Section~\ref{sec:evaluation}, we set this number to 52,445, roughly matching \citet{taori-etal-2023-stanford-alpaca}.

\paragraph{Instance Generation}

To obtain a task $T = (I, S)$ for each generated instruction $I \in \mathcal{I}_{<i}$, we mostly follow \citet{wang-etal-2023-self-instruct}, using the LLM to first identify the task type for~$I$ and then generating the corresponding instances~$S$ based on the task type. Duplicates in $S$ and instances with the same input but different outputs are filtered out. 

Beyond the two task types {\it generation} and {\it classification} covered by \citet{wang-etal-2023-self-instruct}, we further distinguish {\it regression} tasks to enable more fine-grained evaluation with type-specific metrics. We identify the task types for each generated $I$ using templated prompts with instructions sampled from $\mathcal{I}_{0}$ as examples for each task type. Instances are also generated using a templated prompt, providing tasks from $\mathcal{T}_{0}$ representative of each task type. 

\subsection{LLM Instruction Fine-Tuning}

To create our ArgInstruct model, we fine-tune a pretrained LLM on both the entire CA task pool and general tasks, aiming to specialize in CA while maintaining the generalization idea of instruction fine-tuning.
We format task instances into a prompting template for training and mask input tokens during cross-entropy loss calculation, focusing solely on the generated output tokens to help the model retain pre-trained input interpretation skills while ensuring accurate output generation (i.e., mapping from input to output).

\section{Data}
\label{sec:data}
This section first details the selection process for CA seed datasets and tasks that serve as the basis for ArgInstruct. Then, we present the resulting CA-specific instruction fine-tuning dataset, which integrates the seed tasks and newly generated tasks.

\begin{table*}[t]
\small
\renewcommand{\arraystretch}{1}
\setlength{\tabcolsep}{2pt}
\centering
\begin{tabular}{llll@{\hspace*{-2em}}r}
\toprule
& \bf Source                  	                    & \bf Text Genre                    & \bf Tasks Covered by the Dataset                                            & \bf \# Tasks \\
\midrule		
\multirow{10}{*}{\rotatebox{90}{\bf Argument Mining}}
& \newcite{boltuzic-snajder-2014-back}	            & online user comments              &   relation type classification                                              & 1	    \\
& \newcite{peldszus-2015-annotated}	                & short argumentative texts         &   claim extraction, relation identific., function classific.        & 4  	       \\
& \newcite{stab-gurevych-2017-parsing}$^\star$	    & student essays                    &	argumentative span/relation identific., stance classific.       & 4          \\
& \newcite{habernal-gurevych-2017-argumentation}    & user-generated web content        &   persuasiveness detection, toulmin component extract. & 2	           \\
& \newcite{stab-etal-2018-cross}	                & diverse web documents             &   supporting/opposing argument detection                                 & 1	          \\
& \newcite{reimers-etal-2019-classification}	    & web crawl sentences               &   argument similarity prediction                                            & 1	      \\ 
& \newcite{poudyal-etal-2020-echr}	                & court decisions                   &   clause/premise/conclusion recognition, relation predict.           & 4  	        \\
& \newcite{hautli-janisz-etal-2022-qt30}$^\star$	& broadcast political debates       &   propositional/illocutionary relation identification & 2  	        \\
& \newcite{chen-etal-2022-argument}	                & amazon reviews                    &   unit segmentation/classific., helpfulness/relation predict.  & 4	        \\
& \newcite{kuznetsov-etal-2022-revise}$^\star$	    & peer reviews                      &   pragmatic category tagging & 1    	       \\
\midrule
\multirow{10}{*}{\rotatebox{90}{\bf Argument Assessment}}
& \newcite{persing-ng-2015-modeling}	            & student essays                    &   argument strength prediction & 1 	        \\ 
& \newcite{habernal-gurevych-2016-makes}	        & online debates                    &   reason for convincingness prediction & 18  	        \\ 
& \newcite{abbott-etal-2016-internet}$^\star$	    & online debates                    &   agreement/attack/emotion/hostility/sarcasm prediction & 5   \\ 
& \newcite{wachsmuth-etal-2017-computational}	    & online debates                    &   argument quality rating, argumentativeness detection & 15  	        \\ 
& \newcite{habernal-etal-2018-argument}	            & newspaper editorials              &   warrant selection (argument reasoning comprehension) & 1 	        \\ 
& \newcite{gretz-etal-2020-large}$^\star$	        & crowd-sourced arguments           &   argument quality rating, stance prediction & 2  	        \\ 
& \newcite{friedman-etal-2021-overview}	            & crowd-sourced arguments           &   key point generation/matching  & 2	        \\ 
& \newcite{stein-etal-2021-same}	                & online debates                    &   same side stance classification & 1  	        \\ 
& \newcite{heinisch-etal-2022-overview}	            & online political debates          &   (relative) novelty/validity classification & 4	        \\ 
& \newcite{ziegenbein-etal-2023-modeling}	        & reviews, Q\&A, debates            &   inappropriateness (reason) classification  & 14	        \\ 
\midrule
\multirow{10}{*}{\rotatebox{90}{\bf Argument Generation}}
& \newcite{hasan-ng-2014-taking}	                & ideological online debates        &   reason identification & 1  	    \\ 
& \newcite{skeppstedt-etal-2018-less}	            & short argumentative texts         &   argument generation  	    & 1\\ 
& \newcite{wachsmuth-etal-2018-argumentation}$^\star$ & short argumentative texts       &   argument synthesis & 1  	    \\ 
& \newcite{wachsmuth-etal-2018-retrieval}	        & online debates                    &   counter argument generation  	& 4\\ 
& \newcite{roush-balaji-2020-debatesum}$^\star$	    & competitive formal debates        &   extractive debate summarization & 1  	    \\ 
& \newcite{schiller-etal-2021-aspect}$^\star$	    & diverse web documents             &   aspect-based generation & 1  	    \\ 
& \newcite{skitalinskaya-etal-2021-learning}	    & online debates                    &   suboptimal claim detection/improvement & 4  	 \\ 
& \newcite{syed-etal-2021-generating}$^\star$	    & online debates                    &   conclusion generation & 1  	    \\ 
& \newcite{alshomary-etal-2021-belief}	            & online debates                    &   belief-based generation, stance prediction & 2  	    \\ 
& \newcite{stahl-etal-2023-mind}	                & learner essays                    &   enthymeme reconstruction, enthymeme detection & 2  	\\ 
\bottomrule
\end{tabular}
\caption{Overview of the 30 selected CA seed datasets, categorized into argument mining, argument assessment, and argument generation. The table includes the corresponding paper, text genre, and the kinds and numbers of extracted CA seed tasks. The tasks from the 9 CA datasets marked with ``$\star$'' are reserved for testing.} 
\label{tab:dataset-overview}
\end{table*}

\subsection{Task Generation using ArgInstruct}
\label{sec:data-datasets}

\paragraph{CA Seed Dataset Selection}

Initially, we manually collected 71 CA datasets from CA literature. We started from datasets used in shared tasks of the Argument Mining workshop series (6), 
and extended them based on CA papers from the survey of \citet{lauscher-etal-2022-scientia} (30), 
as well as more recent datasets found through searches in 
the ACL Anthology and Google Scholar (35). We then categorized each dataset into one or more CA areas based on their usage in the literature: argument mining, argument assessment, and argument generation.\footnote{We categorized datasets containing tasks from multiple subareas based on their main focus and then the task modeling.}
For each subarea, we selected a subset of ten datasets, consisting of the five most cited datasets along with five additional, lesser-known datasets that cover a diverse range of data sources. This process ensures comprehensive coverage of CA and results in the 30 seed datasets listed in Table~\ref{tab:dataset-overview}. The list of all 71 datasets can be found in Appendix~\ref{app:all-ca-datasets}.

\paragraph{CA Seed Task Collection}

Given the seed data, we obtained the set of CA seed tasks, $\mathcal{T}_0$, from the respective papers. Following our task definition, we consider a task $T = (S, I)$ to model a relation $S$ between inputs and outputs, which can be described in natural language in the form of an instruction $I$. Intermediate steps (e.g., feature extraction or data preprocessing) are not seen as individual CA tasks. If papers contained multiple tasks (e.g., mining and relation identification), each task was treated separately. We wrote the corresponding instructions based on annotation guidelines or, when unavailable, the task descriptions in the papers. To ensure tasks are self-contained, we included relevant term definitions (e.g., class definitions) in the instructions (examples in Appendix~\ref{app:sample-instructions}). In total, we obtained a set of 105 seed tasks, $\mathcal{T}_0 = \{T_1, \ldots, T_{105}\}$, from the 30 seed datasets.

We categorized all see tasks into three task types: \emph{classification}, \emph{regression}, and \emph{generation}. There is a noteworthy connection between task types and CA subareas: argument mining tasks are typically categorized as classification, argument assessment tasks as regression, and argument generation tasks as generation.
However, exceptions exist; for instance, argument mining can be modeled as a classification task at the sentence level or as a generation task at the span level.

\paragraph{CA Task Generation}

Based on the 105 seed tasks, we generated 52,445 additional CA tasks. By roughly matching the size of the general dataset of \citet{taori-etal-2023-stanford-alpaca}, we ensure comparability to their results. Concretely, we used \emph{Meta-Llama-3-70B} as the LLM to generate new CA instructions (step~1 in Figure~\ref{approach}), filter them by CA relevance (step~2), and generate instances (step~3).

\subsection{The ArgInstruct Dataset}

\begin{table*}[t]
\centering
\small
\renewcommand{\arraystretch}{1}
\setlength{\tabcolsep}{2.75pt}
\begin{tabular}{lrrrrrrrrrrrrr}
    \toprule
                        & \multicolumn{4}{c}{\bf \# Instructions}                 &   & \multicolumn{4}{c}{\bf \# Instances}                      &   & \multicolumn{3}{c}{\bf Average Length}      \\
                        \cmidrule{2-5}                                                \cmidrule{7-10}                                                   \cmidrule{12-14}
                        & \bf Classif.    & \bf Regr.    & \bf Generat.    & \bf Total     &   & \bf Classif.    & \bf Regr.    & \bf Generat.    & \bf Total       &   & \bf Instruct.        & \bf Input     & \bf Output    \\
    \midrule
    Seed tasks      & 60          & 23          & 22          & 105           &   & 3,997,517   & 86,870      & 465,652     & 4,550,039       &   & 48.08             & 64.28         & 7.70          \\
    Generated tasks      & 30,204      & 2,170       & 20,071      & 52,445        &   & 30,204      & 2,170       & 20,071      & 52,445          &   & 28.17             & 50.74         & 25.21         \\       
    \bottomrule
\end{tabular}
\caption{Number of instructions and instances in total and per task type (classification, regression, generation), and the average length in words of instructions, non-empty inputs, and outputs for our CA seed and generated tasks.}
\label{tab:training-data-stats}
\end{table*}


Table~\ref{tab:training-data-stats}  compares the statistics for the seed tasks and the generated CA tasks. It can be seen that the seed tasks cover rather few instructions (105) but many instances (4.5M), while the generated tasks have many diverse instructions (52,445) with a single instance each. Classification (30,204) and generation (20,071) dominate the generated tasks, with fewer regression tasks (2,170), likely due to LLMs' limited exposure to regression tasks. Although the generated instructions are shorter on average (28.2 vs.\ 48.1 words), input lengths are similar (50.7 vs.\ 64.3). The longer output length (25.2 vs.\ 7.7) in the generated data likely stems from the higher proportion of generation tasks.

\paragraph{Specialization}

\begin{figure}
    \centering
    \includegraphics[width=0.8\columnwidth]{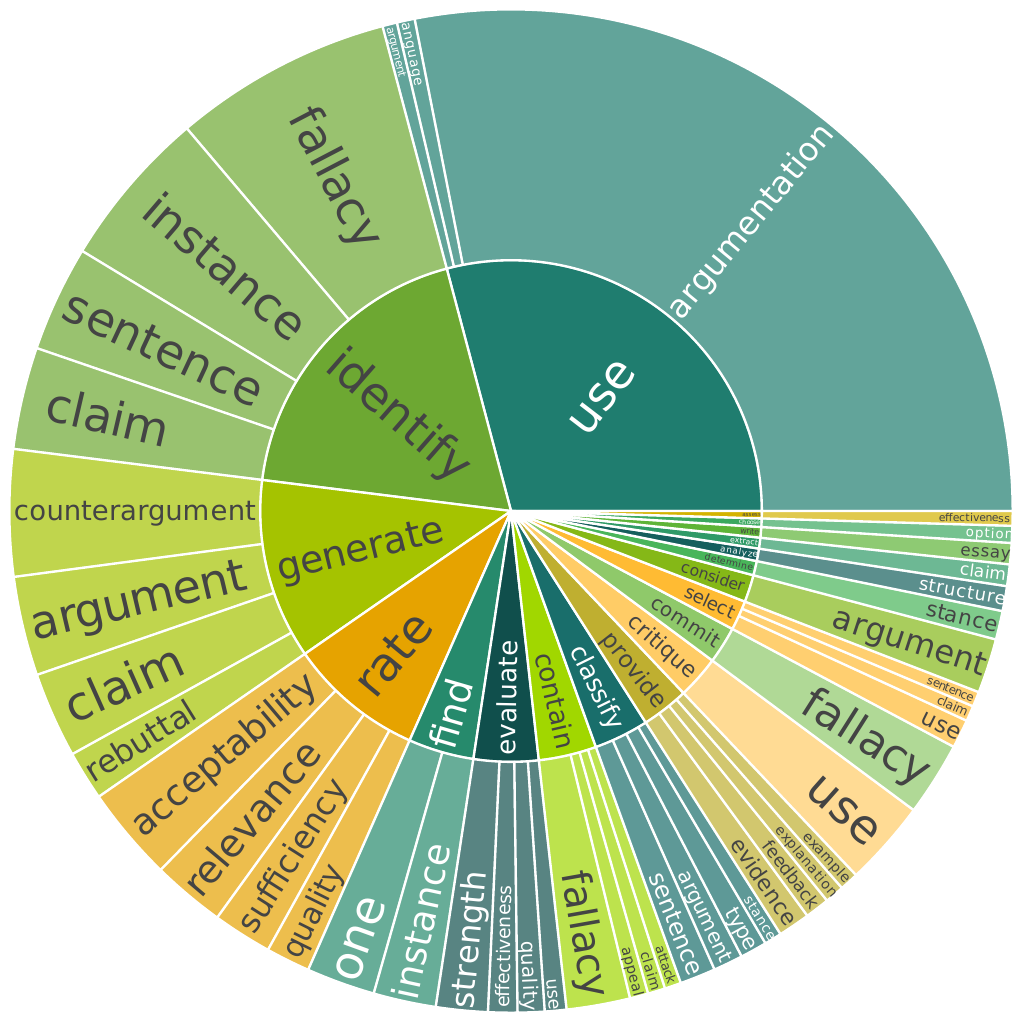}
    \caption{The 20 most common root verbs (inner circle) and their top four direct noun objects (outer circle) in our generated instructions highlight their CA focus.} 
    \label{fig:verb-noun-distribution}
    \hfill
\end{figure}

To assess whether the generated and filtered instructions are indeed CA-specific, we follow \citet{wang-etal-2023-self-instruct} and extract the root verb and its first direct noun object for each generated instruction using the Berkeley Neural Parser \cite{kitaev-etal-2019-multilingual}. Figure~\ref{fig:verb-noun-distribution} shows the 20 most common root verbs and their four most common direct noun objects, which constitute 14\% of all generated instructions. Overall, we see that most direct noun objects are indeed argumentation-related, e.g., ``argumentation'', ``(counter-)argument'' and ``claim''.

\paragraph{Diversity}

To analyze the diversity of the generated instructions, we compute the ROUGE-L F$_1$ similarity between each generated and seed instruction. The average similarity to the closest seed instruction is 0.28, with a maximum of 0.70, which was the similarity threshold~$\tau$ taken from \citet{wang-etal-2023-self-instruct}. Examples of generated instructions within the 10\% closest to seed instructions are:%
%
\begin{itemize}
    \setlength\itemsep{-0pt}
    \item[$I_1$:] ``Extract the central claim from the following argumentative text and predict its stance (pro or con) with respect to the given topic.''
    \item[$I_2$:] ``Determine which of these statements is true: comment 1 attacks argument 2. comment 1 supports argument 2. comment 1 makes no use of argument 2.''
\end{itemize}

While these instructions are close to existing CA tasks, namely thesis extraction and stance detection ($I_1$), and relation detection ($I_2$), they introduce new wordings that will likely lead to more robust fine-tuning. Exemplary instructions with the lowest maximal similarity to the seed instructions are:%
\begin{itemize}
    \setlength\itemsep{-0pt}
    \item[$I_3$:]  ``For each elementary unit x, choose one proposition y such that there exists a support relation between x and y.''
    \item[$I_4$:]  ``Generate a list of possible explanations for why someone might believe something based on their background and experiences.''    
\end{itemize}

$I_3$ is a form of argumentative relation extraction, while $I_4$ is an argument generation task that, to our knowledge, has not yet been studied. The final task involves identifying various fallacies, which are related to CA but were not covered by the seed tasks. For more examples, see Appendix~\ref{app:sample-generated-instructions}.

\paragraph{Quality}
To evaluate the quality of the generated tasks, two authors of this paper manually evaluated a random sample of 200 generated tasks. The annotation protocol was adapted from the questionnaire of \citet{wang-etal-2023-self-instruct}, assessing three key aspects:
\begin{itemize}
    \setlength\itemsep{-0pt}
    \item[$Q_1$:] Does the instruction describe a valid CA task? 
    \item[$Q_2$:] Is the input appropriate for the instruction?
    \item[$Q_3$:] Is the output a correct and acceptable response to the instruction and input? 
\end{itemize}

The tendencies of our results are similar to those reported by \citet{wang-etal-2023-self-instruct}, with 87\% of \nopagebreak{the instructions} corresponding to valid CA tasks ($Q_1$), 75.5\% of the input instances deemed appropriate for their instructions ($Q_2$), and 62.5\% of the outputs satisfying all correctness criteria ($Q_3$). Hence, we conclude that, although the generated dataset contains some noise, most of the generated tasks are entirely or at least partially correct.
\section{Evaluation}
\label{sec:evaluation}

\newcommand{\fullCircle}{\begin{tikzpicture}
    \filldraw[black] (0,0) circle (0.8ex);
\end{tikzpicture}}

\newcommand{\fullCircleGray}{\begin{tikzpicture}
    \fill[gray!40] (0,0) circle (0.8ex);
    \draw[black] (0,0) circle (0.8ex);
\end{tikzpicture}}

\newcommand{\halfCircle}{\begin{tikzpicture}
    \fill[black] (0,0) -- (90:0.8ex) arc (90:-90:0.8ex) -- cycle;
    \draw (0,0) circle (0.8ex);
\end{tikzpicture}}

\newcommand{\halfCircleGray}{\begin{tikzpicture}
    \fill[black] (0,0) -- (90:0.8ex) arc (90:-90:0.8ex) -- cycle;
    \fill[gray!40] (0,0) -- (90:0.8ex) arc (90:270:0.8ex) -- cycle;
    \draw (0,0) circle (0.8ex);
\end{tikzpicture}}

\newcommand{\thirdCircle}{\begin{tikzpicture}
    \fill[black] (0,0) -- (90:0.8ex) arc (90:-30:0.8ex) -- cycle;
    \draw (0,0) circle (0.8ex);
\end{tikzpicture}}

\newcommand{\thirdCircleGray}{\begin{tikzpicture}
    \fill[gray!40] (0,0) circle (0.8ex);
    \fill[black] (0,0) -- (90:0.8ex) arc (90:-30:0.8ex) -- cycle;
    \draw (0,0) circle (0.8ex);
\end{tikzpicture}}

\newcommand{\emptyCircle}{\begin{tikzpicture}
    \draw[black] (0,0) circle (0.8ex);
\end{tikzpicture}}

\begin{table*}[t]
	\centering
	\small
	\renewcommand{\arraystretch}{1}
	\setlength{\tabcolsep}{2pt}
	\begin{tabular}{lccclrrrrrrrrr}
		\toprule
                                        & \multicolumn{3}{c}{\bf Fine-Tuning Data}                     && \multicolumn{4}{c}{\bf (a) Unseen CA Instances}                                        && \multicolumn{4}{c}{\bf (b) Unseen CA Tasks} \\
                                        \cmidrule{2-4}                                              \cmidrule{6-9}                                                                               \cmidrule{11-14}
        \bf Approach                    & \bf seedCA        & \bf genCA         & \bf general       && \bf F$_1$ $\uparrow$ & \bf MASE$\downarrow$ & \bf R-L  $\uparrow$ & \bf Rank$\downarrow$  & & \bf F$_1$ $\uparrow$ & \bf MASE$\downarrow$ & \bf R-L  $\uparrow$ & \bf Rank$\downarrow$\\
        \midrule
        Gemma-2-9B (baseline)           & \emptyCircle{}    & \emptyCircle{}    & \emptyCircle{}    && .45                  & 1.6                   & .39                 &  9.0\phantom{a}                & & .45\phantom{a}                  & 4.2\phantom{aa}                        & .15\phantom{a}                 & 10.7                     \\  
         + seedCA                       & \fullCircle{}     & \emptyCircle{}    & \emptyCircle{}    && \bf \underline{.65}  & 1.1                   & \bf \underline{.50} &  \bf \underline{2.3}\phantom{a} & & \bf \underline{.65}\phantom{a} & 2.5\phantom{aa}                        & .23\phantom{a}                 &  5.7                    \\  
         + genCA                        & \emptyCircle{}    & \fullCircle{}     & \emptyCircle{}    && .51                  & 2.7                   & .45                 &  8.7\phantom{a}                & & .52\phantom{a}                  & 3.0\phantom{aa}                        & .30\phantom{a}                 &  6.3                    \\  
         + general                      & \emptyCircle{}    & \emptyCircle{}    & \fullCircle{}     && .50                  & 2.1                   & .32                 & 10.0\phantom{a}                & & .51\phantom{a}                  & 2.6\phantom{aa}                        & .29\phantom{a}                 &  6.7                    \\  
         + seedCA, genCA                & \halfCircle{}     & \halfCircle{}     & \emptyCircle{}    && .61                  & 1.6                   & .49                 &  4.0\phantom{a}                & & .57\phantom{a}                  & 2.9\phantom{aa}                        & .26\phantom{a}                 &  5.7                    \\ 
         + genCA, general               & \emptyCircle{}    & \halfCircle{}     & \halfCircle{}     && .51                  & 1.7                   & .39                 &  8.7\phantom{a}                & & .50\phantom{a}                  & 3.0\phantom{aa}                        & .17\phantom{a}                 & 8.3                    \\  
         + seedCA, general              & \halfCircle{}     & \emptyCircle{}    & \halfCircle{}     && .59                  & \bf \underline{1.0}   & .48                 &  4.0\phantom{a}                & & .60\phantom{a}                  & \underline{2.3}\phantom{aa}            & .30\phantom{a}                 &  4.0                    \\  
         \bf + seedCA, genCA, general   & \thirdCircle{}    & \thirdCircle{}    & \thirdCircle{}    && .61                  & 1.5                   & .49                 &  4.3\phantom{a}                & & \bf \underline{.65}\phantom{a}  & 2.5\phantom{aa}                        & \bf \underline{.32}\phantom{a} & \bf \underline{2.0}         \\  

        \addlinespace       
        Gemma-2-9B-General (baseline)   & \emptyCircle{}    & \emptyCircle{}    & \fullCircleGray{} && .48                  & 2.3                   & .34                 & 11.0\phantom{a}                & & .50\phantom{a}                  & 2.1\phantom{aa}                        & .24\phantom{a}                 & 7.0                    \\  
         + seedCA                       & \fullCircle{}     & \emptyCircle{}    & \fullCircleGray{} && \underline{.64}      & \underline{1.2}       & \underline{.49}     & \underline{2.7}\phantom{a}      & & .63\phantom{a}                  & 2.1\phantom{aa}                        & \bf \underline{.32}\phantom{a} & 4.3                    \\    
         + genCA                        & \emptyCircle{}    & \fullCircle{}     & \fullCircleGray{} && .49                  & 2.6                   & .44                 &  9.3\phantom{a}                & & .52\phantom{a}                  & 2.6\phantom{aa}                        & .30\phantom{a}                 & 6.7                    \\
         \bf + seedCA, genCA (ArgInstruct)  & \halfCircle{} & \halfCircle{}     & \fullCircleGray{} && .57                  & 1.3                   & \underline{.49}     &  4.0\phantom{a}                & & \bf \underline{.65}$^\dagger$    & \bf \underline{1.9}$^{\dagger\ddagger}$ & .31$^\dagger$                  & \bf \underline{2.0}    \\ 
         
		\bottomrule
	\end{tabular}
    \caption{Main CA results on (a) unseen CA instances and (b) unseen CA tasks: 
    Gemma-2-9B instruction fine-tuned on 52k instances of CA seed ({\it +seedCA}), generated CA ({\it +genCA}), general ({\it +general}) tasks and their combinations. The symbols represent the proportion of fine-tuning data coming from each source (\texttt{\protect\emptyCircle}: 0\%, \texttt{\protect\thirdCircle}: 33\%, \texttt{\protect\halfCircle}: 50\%, \texttt{\protect\fullCircle}: 100\%). \texttt{\protect\fullCircleGray} indicates prior use for general instruction fine-tuning. The best values are bold, the best per base model underlined. Overall, both full specialized instruction fine-tuning variants (bold) achieve the best mean \emph{rank}. $\dagger$ and $\ddagger$ denote significant improvements over the baseline and {\it +seedCA} respectively (Wilcoxon signed-rank test, $p<.05$).}
	\label{tab:generated-existing-evaluation}
\end{table*}

We now present our experiments to evaluate the impact of specialized instruction fine-tuning. LLM variants were trained on (a) CA seed tasks, (b) generated CA tasks, (c) general tasks, and (d) combinations thereof.  We then assessed their performance on unseen instances of known CA tasks and their generalizability to entirely unseen CA tasks. 

\subsection{Experimental Setup}

\paragraph{Data}
We reserved 21 seed tasks (20\% of 105) from nine CA datasets as unseen test tasks (Table~\ref{tab:dataset-overview}), balancing across argument mining, assessment, and generation. The remaining 84 seed tasks were split into training, validation, and test sets, using prior splits where available or a 7:1:2 random split otherwise. For evaluation, we sampled 100 test instances per task, balancing labels for classification and covering the full range for regression, while generation tasks were sampled randomly. The same sampling method was used to assemble 52,445 instances from the training seed tasks for our (\emph{seedCA}) dataset.

\begin{table}[t]
	\centering
	\small
	\renewcommand{\arraystretch}{1}
	\setlength{\tabcolsep}{3pt}
	\begin{tabular}{llcccc}
		\toprule
        \bf Approach                         && \bf F$_1$ $\uparrow$ & \bf R-L $\uparrow$ & \bf Rank $\downarrow$\\              
        \midrule
            Gemma-2-9B                    && .55                  & \bf .43                & 3.0                     \\
            \bf + seedCA, genCA, general      && .61                  & .40                    & 3.5                     \\
            \addlinespace
            Gemma-2-9B-General            && \bf .62              & .37                    & 2.5                     \\ 
            \bf + seedCA, genCA (ArgInstruct) && \bf .62              & \bf .43                & \bf 1.0                 \\
        \bottomrule
	\end{tabular}
	\caption{Generalization results on SuperNI: Zero-shot performance of \emph{Gemma-2-9B}, its instruction fine-tuned variant \emph{Gemma-2-9B-General}, and our CA-specialized LLMs. \emph{ArgInstruct} performs best across all metrics.}
	\label{tab:superni-evaluation}
\end{table}
\begin{table*}[h]
    \centering
    \small
    \renewcommand{\arraystretch}{1}
    \setlength{\tabcolsep}{1.5pt}
    \begin{tabular}{lll@{}rcl}
        \toprule
        \bf Area	& \bf Task (Source)                                                 & \bf Unseen             & \bf Metric           & \bf ArgInstruct & \bf Task-Specific SOTA                            \\
        \midrule
        Mining		& Argument Detection \& Classif. \cite{stab-etal-2018-cross}          & Instances            & F$_1$ $\uparrow$     & .62             & {.73} \cite{wang-etal-2024-target-aware}      \\
       				& Relation Detection \cite{stab-gurevych-2017-parsing}              & Task                 & F$_1$ $\uparrow$     & .47             & {.84} \cite{cabessa-etal-2025-argument}      \\
        Assessment 	& Inappropriateness Classif. \cite{ziegenbein-etal-2023-modeling}   & Instances            & F$_1$ $\uparrow$     & .74             & {.75} \cite{ziegenbein-etal-2023-modeling}  \\
        				& Argument Quality Rating \cite{gretz-etal-2020-large}              & Task                 & MAE $\downarrow$     & .25             & {.13} \cite{bao-etal-2024-comparison}        \\
        Generation 	& Enthymeme Reconstruction \cite{stahl-etal-2023-mind}              & Instances            & R-L $\uparrow$       & .16             & {.17} \cite{stahl-etal-2023-mind}            \\
       				&  Argument Summarization \cite{roush-balaji-2020-debatesum}         & Task                 & R-L $\uparrow$       & .56             & {.57} \cite{roush-balaji-2020-debatesum}     \\
        \bottomrule
    \end{tabular}
    \caption{Performance comparison of our ArgInstruct LLM with the state-of-the-art (SOTA) upper bound on six CA seed tasks: three tasks were included in our training data (instances unseen), and three were entirely new to our  LLM (task unseen). In contrast, the SOTA models are trained in a supervised manner on the single task.}
    \label{tab:task-specific-sota}
\end{table*}

\paragraph{Evaluation}

We use guided generation \cite{willard-2023-efficient} for classification and regression tasks and \emph{open} generation (up to $512$ new tokens) for generation tasks. For classification and regression, a finite state machine decodes model outputs for direct comparison with ground-truth values. 

\paragraph{Metrics}

To ensure meaningful comparisons, we use task-specific evaluation metrics. For classification, we report the micro-averaged F$_1$-score on balanced test sets. For regression, we use mean absolute scaled error (MASE), which normalizes errors with respect to a na\"ive mean baseline, with scores less than $1$ indicating better performance and above $1$ indicating worse \cite{hyndman-athanasopoulos-2021}. For generation, we report the ROUGE-L F$_1$-score. Additionally, we calculate the mean rank for each model, reflecting its performance across datasets and tasks, providing a comprehensive measure of overall CA performance.

\paragraph{Models}

As base models, we use the recent LLM \emph{Gemma-2-9B} \cite{gemma-team-etal-2024-gemma} and an instruction fine-tuned variant, \emph{Gemma-2-9B-General}. The latter is obtained by fine-tuning \emph{Gemma-2-9B} on the general instruction dataset of \citet{taori-etal-2023-stanford-alpaca}, ensuring that it has not been exposed to the SuperNI benchmark for general instruction following \cite{wang-etal-2022-super}. Both models are then instruction fine-tuned on equally-sized combinations of seed CA tasks, generated CA tasks, and general tasks (52k instances each). Details on hyperparameter tuning can be found in Appendix~\ref{app:hpo-details}. 

\subsection{Results of ArgInstruct on CA Tasks}

Table~\ref{tab:generated-existing-evaluation} reports the CA performance of the two base LLMs and the LLMs trained with \emph{ArgInstruct} and its ablations, averaged across datasets. We evaluate performance on (a) unseen test instances of the training tasks and (b) test instances of the completely unseen test tasks. Note that for the base models, all instances and tasks are unseen. 

On unseen instances, LLMs fine-tuned on the respective training tasks (\emph{+seedCA}) work best (ranks $2.3$ and $2.7$), as expected. For unseen tasks, combining \emph{seedCA}, \emph{genCA}, and \emph{general} performs best with either base model (both rank $2.0$), outperforming ablated variants. This strongly supports our hypothesis that specialized instruction fine-tuning enhances generalization on CA tasks. Whether general instruction fine-tuning occurs prior (\emph{Gemma-2-9B-General+seedCA,genCA}) or alongside CA-specific fine-tuning (\emph{Gemma-2-9B+seedCA,genCA,general}) has little effect on performance, although the latter performs slightly better on unseen instances.
Appendix~\ref{app:full-results-table} reports the results of the latter variant on all 105 tasks.

\subsection{Results of ArgInstruct on General Tasks}

To assess if specialized instruction fine-tuning preserves generalization capabilities, Table~\ref{tab:superni-evaluation} shows the performance of both base models and the full CA-specialized LLMs on the SuperNI benchmark of general NLP tasks \cite{wang-etal-2022-super}. Unlike above, the timing of general instruction fine-tuning has an effect here: Our specialized LLM, fine-tuned on general instructions first (\emph{Gemma-2-9B-General+seedCA,genCA}), performs best across all metrics. Given its strong results on both CA and general tasks, we designate it as our final model, now referred to as the \emph{ArgInstruct} model.

\subsection{Comparison to Task-Specific Fine-Tuning}

To assess the actual strength of our \emph{ArgInstruct} model, we compare it on specific CA tasks against the current state-of-the-art (SOTA) on the entire original test sets of six tasks -- two each from argument mining, assessment, and generation (one seen during training and one entirely unseen). This reveals the trade-off between specializing in CA as a whole and developing task-specific approaches. 

As shown in Table~\ref{tab:task-specific-sota}, the SOTA approaches win on all six tasks, but ArgInstruct achieves comparable performance in three of them. We speculate that the performance gap arises because the SOTA models (a) benefit from a larger number of task-specific training instances and/or (b) are able to better adjust to the single task and data source. Nonetheless, \emph{ArgInstruct} is a strong and versatile model, offering broad generalization across CA tasks in a zero-shot setting. However, depending on the CA task, additional task-specific fine-tuning may further enhance the performance of \emph{ArgInstruct} for optimal results.

\subsection{Comparison to General LLMs}
\label{sec:otherllms}

To understand the instruction-following abilities of our base model, we finally compare it against competitive instruction-following LLMs of similar size \cite{taori-etal-2023-stanford-alpaca, jiang-2023-mistral7b, gemma-team-etal-2024-gemma, grattafiori-2024-llama3, openai-2024-gpt4ocard}. 
Table~\ref{tab:benchmark-evaluation} shows the zero-shot performance of all models alongside \emph{Majority} and \emph{Random} baselines. ArgInstruct outperforms all others in terms of F$_1$~(.65). However, for regression tasks, no model proves reliable, as all MASE scores are worse than predicting the mean. In generation (R-L), \emph{GPT-4o-mini} appears slightly superior to our model (.32 vs. .31), though its comparability may be limited due to its unknown size. Overall, \emph{ArgInstruct} achieves the strongest result on unseen CA tasks, achieving the best mean rank ($2.33$) across all models.


\begin{table}[t]
	\centering
	\small
	\renewcommand{\arraystretch}{1}
	\setlength{\tabcolsep}{4pt}
	\begin{tabular}{llcccc}
		\toprule
        \bf Model                   && \bf F$_1$ $\uparrow$ & \bf MASE $\downarrow$ & \bf R-L  $\uparrow$ & \bf Rank $\downarrow$\\    
        \midrule
        Majority                    && .38                  & \bf 1.2               & .18                 & 6.33                 \\ 
        Random                      && .34                  & 1.4                   & .17                 & 6.33                 \\  
        \addlinespace
        Alpaca-7B-it                && .44                  & 2.3                   & .18                 & 5.67                 \\  
        Gemma-2-9B-it               && .62                  & 3.0                   & .22                 & 5.33                 \\ 
        LLaMA-3-8B-it               && .48                  & 2.5                   & .22                 & 6.00                 \\ 
        Ministral-8B-it             && .50                  & 2.6                   & .24                 & 5.00                 \\
        Mistral-7B-it               && .61                  & 2.1                   & .26                 & 3.33                 \\ 
        GPT-4o-mini                 && .59                  & 1.5                   & \bf .32             & 2.67                 \\ 
        \addlinespace
        ArgInstruct (Ours)          && \bf .65              & 1.9                   & .31                 & \bf 2.33             \\     
	\bottomrule
	\end{tabular}
	\caption{Zero-shot evaluation of our \emph{ArgInstruct} model compared to recent instruction fine-tuned models of similar size (besides GPT-4o-mini), on our CA test tasks.}
	\label{tab:benchmark-evaluation}
\end{table}

\section{Conclusion}
Despite their strong generalization capabilities, instruction-following LLMs 
struggle with tasks that require domain knowledge. 
We propose \emph{Arg\-Instruct}, a new specialized instruction fine-tuning method to address this issue for the domain of computational argumentation (CA). 

As a starting point, we have collected 105 CA tasks from the literature and crafted natural instructions for each that serve as a benchmark for LLM-based CA. Additionally, we have generated 52k CA-specific tasks, adapting the self-instruct process to bridge between generalization and CA specialization. We have then trained CA-specialized instruction-following LLMs, combining the collected and generated CA tasks with general instruction fine-tuning data. Our experiments suggest that an LLM fine-tuned on the combined data performs best on unseen CA tasks without losing its general instruction-following capabilities. While the LLM did not fully reach single-task SOTA results, it is on par in half the tasks. At the same time, it outperforms several existing instruction-following models, including the proprietary GPT-4o-mini.

We conclude that our \emph{ArgInstruct} method denotes a substantial step towards overcoming the domain challenges of LLMs, providing a benchmark dataset and a task-agnostic model for CA. We expect that our method may be well-transferable to other specialized NLP domains, for example, to the educational domain. There, our method could involve collecting seed tasks such as essay scoring, feedback generation, text suggestion, and rewriting, but we leave this to future work.
\section{Limitations}

The research proposed in this paper may have limitations with respect to four aspects that we discuss in the following: (1) Model training, (2) model evaluation, (3) quality of the generated data, and (4) generalizability to other domains.

\paragraph{Model Training} 
We only use a subset of training instances for the training of our models. While this is a strength of our method in that we only require $\approx$ 500 instances per task, using all training instances could further increase the performance of models, and a different sampling of instances may lead to slightly different results also depending on the varying quality of training data instances.

\paragraph{Model Evaluation}  
We point to the common problem of evaluating generative models with automatic metrics (here, ROUGE-L). Beyond related research, we at least used different measures to give adequate insights into task-related differences. In addition, we decided on a balanced evaluation of 100 instances per task to achieve a uniform, systematic setting. This is, of course, only an approximation of evaluating the full datasets, such that some values could change in tasks that are heavily instance-dependent. While our comparison of the models mentioned in this paper is fair, we do not recommend taking these values blindly and comparing them directly to approaches from related work that are evaluated on full test sets. However, we hope that our analysis in Section \ref{sec:data}, which uses the entire test sets, gives readers an idea about the transferability. Finally, given the generally poor performance of all LLMs on regression tasks, we do not recommend using them for such tasks.

\paragraph{Quality of Generated Data} 
Although we did our best to select a representative subset of datasets, we covered only a subset of the CA datasets found during our literature search (30 out of 71). The reason for this is the high manual effort required to manually craft the instructions and load and parse the respective datasets. However, we believe that our methodology would benefit from having more of these existing datasets, as it potentially decreases the amount of data that needs to be generated and could lead to an increased quality of the data in terms of diversity and instance quality. While a manual examination of the generated data, the statistics provided in the paper, and the manual evaluation of the self-instruct method by \citet{wang-etal-2023-self-instruct} suggest a good quality of the data, we ultimately do not know the quality of the generated tasks, and whether their instances correctly match their corresponding instructions.

\paragraph{Generalizability} 
The success of the \emph{ArgInstruct} model depends significantly on the availability of a diverse set of tasks to be used for instructions. Its performance may be limited in domains where task data is scarce or difficult to collect, affecting the model’s generalizability. While we only instantiated \emph{ArgInstruct} for computational argumentation, we expect and encourage future work to apply our proposed methodology to other NLP areas that require specific domain knowledge.

\section*{Acknowledgments}
The presented work has been partially funded by the Deutsche Forschungsgemeinschaft (DFG, German Research Foundation) within the project ArgSchool, project number 453073654, and within the project OASiS, project number 455913891, which is part of the Priority Program “Robust Argumentation Machines (RATIO)” (SPP-1999). Furthermore, this work was supported by the Federal Ministry of Education and Research (BMBF), Germany, under the AI service center KISSKI (grant number 01IS22093C).

\bibliography{acl24-argpaca-lit}

\appendix
\section{Collected CA Datasets}
\label{app:all-ca-datasets}

Table~\ref{tab:all-datasets-overview} shows the complete list of the 71 CA datasets considered.

\begin{table*}[t]
\small
\renewcommand{\arraystretch}{1}
\setlength{\tabcolsep}{15pt}
\centering
\begin{tabular}{lll}
    \toprule                    
    \bf Argument Mining	                                    & \bf Argument Assessment                                   & \bf Argument Generation \\      
    \midrule                                                 
    \newcite{al-khatib-etal-2016-news}                      & \newcite{abbott-etal-2016-internet}$^\heartsuit$               & \newcite{alshomary-etal-2021-belief}$^\heartsuit$ \\
    \newcite{al-khatib-etal-2016-cross}                     & \newcite{ajjour-etal-2019-modeling}                       & \newcite{eden-etal-2023-welcome} \\
    \newcite{alhindi-ghosh-2021-sharks}                     & \newcite{beck-etal-2021-investigating}                    & \newcite{hasan-ng-2014-taking}$^\heartsuit$ \\
    \newcite{bar-haim-etal-2017-stance}                     & \newcite{friedman-etal-2021-overview}$^\heartsuit$             & \newcite{jo-etal-2020-detecting} \\ 
    \newcite{boltuzic-snajder-2014-back}$^\heartsuit$            & \newcite{gleize-etal-2019-convinced}                      & \newcite{roush-balaji-2020-debatesum}$^\heartsuit$ \\	             
    \newcite{chen-etal-2022-argument}$^\heartsuit$               & \newcite{gretz-etal-2020-large}$^\heartsuit$                   & \newcite{schiller-etal-2021-aspect}$^\heartsuit$ \\	        
    \newcite{eckle-kohler-etal-2015-role}                   & \newcite{habernal-etal-2018-argument}$^\heartsuit$             & \newcite{skeppstedt-etal-2018-less}$^\heartsuit$  \\        
    \newcite{eindor-etal-2020-corpus}                       & \newcite{habernal-etal-2018-name}                         & \newcite{skitalinskaya-etal-2021-learning}$^\heartsuit$ \\
    \newcite{feger-dietze-2024-taco}                        & \newcite{habernal-gurevych-2016-argument}                 & \newcite{stahl-etal-2023-mind}$^\heartsuit$ \\
    \newcite{grundler-etal-2022-detecting}                  & \newcite{habernal-gurevych-2016-makes}$^\heartsuit$            & \newcite{syed-etal-2021-generating}$^\heartsuit$ \\ 
    \newcite{habernal-gurevych-2017-argumentation}$^\heartsuit$  & \newcite{heinisch-etal-2022-overview}$^\heartsuit$ 	        & \newcite{wachsmuth-etal-2018-argumentation}$^\heartsuit$ \\ 
    \newcite{haddadan-etal-2019-yes}                        & \newcite{persing-etal-2010-modeling}                      & \newcite{wachsmuth-etal-2018-retrieval}$^\heartsuit$\\      
    \newcite{hautli-janisz-etal-2022-qt30}$^\heartsuit$          & \newcite{persing-ng-2013-modeling} \\
    \newcite{hidey-etal-2017-analyzing}                     & \newcite{persing-ng-2014-modeling} \\
    \newcite{kuznetsov-etal-2022-revise}$^\heartsuit$            & \newcite{persing-ng-2015-modeling}$^\heartsuit$ \\
    \newcite{lauscher-etal-2018-argument}                   & \newcite{persing-ng-2016-modeling} \\
    \newcite{liebeck-etal-2016-airport}                     & \newcite{sobhani-etal-2015-argumentation} \\
    \newcite{mayer-etal-2020-transformer}                   & \newcite{stab-gurevych-2017-recognizing} \\	        
    \newcite{ong-etal-2014-ontology}                        & \newcite{stein-etal-2021-same}$^\heartsuit$ \\	             
    \newcite{park-cardie-2018-corpus}                       & \newcite{toledo-etal-2019-automatic}  \\	                     
    \newcite{peldszus-2015-annotated}$^\heartsuit$               & \newcite{vamvas-sennrich-2020-x-stance} \\ 
    \newcite{poudyal-etal-2020-echr}$^\heartsuit$                & \newcite{wachsmuth-etal-2017-computational}$^\heartsuit$ \\
    \newcite{reimers-etal-2019-classification}$^\heartsuit$      & \newcite{walker-etal-2012-corpus} \\
    \newcite{rinott-etal-2015-show}                         & \newcite{ziegenbein-etal-2023-modeling}$^\heartsuit$ \\
    \newcite{schaller-etal-2024-darius} \\                    
    \newcite{shnarch-etal-2018-will} \\                       
    \newcite{shnarch-etal-2020-unsupervised} \\                
    \newcite{stab-etal-2018-cross}$^\heartsuit$ \\	                     
    \newcite{stab-gurevych-2017-parsing}$^\heartsuit$ \\
    \newcite{stahl-etal-2024-school} \\
    \newcite{toledo-ronen-etal-2020-multilingual} \\
    \newcite{trautmann-2020-aspect} \\
    \newcite{trautmann-etal-2020-fine} \\
    \newcite{visser-etal-2019-annotated} \\
    \newcite{wambsganss-etal-2020-corpus} \\
    \bottomrule
\end{tabular}
\caption{The list of all 71 considered CA datasets. The 30 datasets selected as seed datasets for generating CA tasks are marked with ``$\heartsuit$''.} 
\label{tab:all-datasets-overview}
\end{table*}

\section{Exemplary Seed Instructions}
\label{app:sample-instructions}

This section provides examples of manually crafted seed instructions for different tasks and datasets. While listing all 105 instructions here would be impractical, the complete set of seed instructions can be found in the provided code.

\paragraph{Argument Mining}
\begin{itemize}
    \setlength\itemsep{-0pt}
    \item {\bf Argument Detection \& Classification} \cite{stab-etal-2018-cross}: ``Given a sentence and a topic, classify the sentence as a ``supporting argument'' or ``opposing argument'' if it includes a relevant reason for supporting or opposing the topic, or as a ``non-argument'' if it does not include a reason or is not relevant to the topic.''
    \item {\bf Argument Component Classification} \cite{stab-gurevych-2017-parsing}: ``Given the following essay as context, and a list of argumentative components extracted from the essay. Label each argumentative component as ``major claim'', ``claim'', or ``premise''.''    
\end{itemize}

\paragraph{Argument Assessment}
\begin{itemize}
    \setlength\itemsep{-0pt}
    \item {\bf Inappropriateness Detection} \cite{ziegenbein-etal-2023-modeling}: ``An argument is appropriate if the used language supports the creation of credibility and emotions as well as if it is proportional to its topic. Given the following argument and the topic of the debate the argument appeared in. Decide whether the argument is Appropriate or Inappropriate.''
    \item {\bf Overall Quality Rating} \cite{wachsmuth-etal-2017-computational}: ``How would you rate the overall quality of the author’s argumentation on the scale ``1'' (Low), ``2'' (Average) or ``3'' (High)?''
\end{itemize}

\paragraph{Argument Generation}
\begin{itemize}
    \setlength\itemsep{-0pt}
    \item {\bf Enthymeme Reconstruction} \cite{stahl-etal-2023-mind}: ``An enthymeme is defined here as any missing argumentative discourse unit (ADU) that would complete the logic of a written argument. Is there a problematic enthymematic gap at the position marked with ``<mask>'' in the following argument?''
    \item {\bf Argument Summarization} \cite{roush-balaji-2020-debatesum}: ``Create a word-level extractive summary of the argument by ``underlining'' and/or ``highlighting'' the evidence in such a way to support the argument being made.''
\end{itemize}

\section{Exemplary Generated Instructions}
\label{app:sample-generated-instructions}

Examples from the 10\% generated instructions with the highest maximal similarity to the seed instructions are:

\begin{itemize}
    \item[$I_5$:] ``Consider the following arguments (argument a and argument b). Would you agree with the following statement? Argument a has worse reasoning because it presents facts without explaining their relevance to the claim.''
    \item[$I_6$:] ``Given a question, stance (yes vs. no) towards this question and a premise, your task is to form a counterargument against the given stance using the given premise.''
    \item[$I_7$:] ``Given two arguments, determine whether they have the same stance towards their common topic.''
\end{itemize}

Further exemplary instructions from the 10\% generated instructions with the lowest maximal similarity to the seed instructions are:

\begin{itemize}
    \item[$I_8$:] ``Classify the type of logical fallacy committed in the argument (ad hominem, appeal to emotion, appeal to ignorance, appeal to popularity, appeal to tradition, circular reasoning, confirmation bias, false dichotomy, genetic fallacy, post hoc ergo propter hoc, red herring, slippery slope, straw man, tu quoque).''
    \item[$I_9$:] ``This task requires you to identify whether each statement below expresses a subjective opinion or objective fact. please read all instructions carefully before starting! subjective opinions are personal judgments about things that cannot be proven true or false objectively. they reflect individual preferences and values rather than empirical observations or established knowledge. objective facts are verifiable pieces of information derived from observation, experimentation, measurement, calculation, etc… they describe physical reality independent of human perception or interpretation. examples include historical events, scientific laws, mathematical formulas, etc.''
    \item[$I_{10}$:] ``Given a hypothetical scenario wherein there exists conflict over ownership rights pertaining to certain property located within disputed territory between two neighboring countries; construct logical sequence of steps leading up towards resolution through negotiation process involving both parties concerned alongside third party mediator appointed mutually agreed upon basis taking into account all relevant factors such as historical background surrounding issue at hand.''
\end{itemize}

\section{Hyperparameter-Tuning}
\label{app:hpo-details}

We use adapter-based low-rank adaptation (LoRA) \cite{hu-2021-lora} with $r=16$, an amplification factor of $32$, and a dropout rate of $0.05$ to enhance training efficiency. To determine an optimal learning rate, number of epochs, and batch size, we use Optuna \cite{akiba-2019-optuna} for hyperparameter optimization. Ultimately, all models are trained for 7 epochs with a learning rate of $9.88 \times 10^{-5}$, an effective batch size of $64$, cosine learning rate decay, and a warmup ratio of $0.05$.

\section{ArgInstruct: Task Results}
\label{app:full-results-table}

Table~\ref{tab:arginstruct-eval-per-task} and \ref{tab:arginstruct-eval-per-task2} show the performance of our \emph{ArgInstruct} model for the 100 sampled test instances for all 105 CA seed tasks.

\begin{table*}[t]
    \centering
    \small
    \renewcommand{\arraystretch}{1}
    \setlength{\tabcolsep}{5pt}
    \begin{tabular}{llllccc}
        \toprule
        \bf Source & \bf Task & \bf Split & \bf F$_1$ $\uparrow$ & \bf MASE $\downarrow$ & \bf R-L $\uparrow$ \\
        \midrule
        \citet{abbott-etal-2016-internet} & Predict Agreement & test & - & 2.28 & - \\ 
        & Predict Respect & test & - & 2.09 & - \\ 
        & Predict Factuality & test & - & 1.94 & - \\ 
        & Predict Nice & test & - & 2.16 & - \\ 
        & Predict Sarcasm & test & - & 3.15 & - \\
        \midrule
        \citet{alshomary-etal-2021-belief} & Stance Prediction & train & 0.42 & - & - \\ 
        & Belief Based Claim Generation & train & - & - & 0.13\\
        \midrule
        \citet{boltuzic-snajder-2014-back} & Stance Detection & train & 0.39 & - & - \\
        \midrule
        \citet{chen-etal-2022-argument} & Review Helpfulness Prediction & train & - & 1.02 & - \\ 
        & Relation Detection & train & 0.63 & - & - \\ 
        & Unit Segmentation Prediction & train & - & - & 0.95\\ 
        & Init Classification Prediction & train & - & - & 0.69\\
        \midrule
        \citet{friedman-etal-2021-overview} & Key Point Matching & train & 0.76 & - & - \\ 
        & Key Point Generation & train & - & - & 0.31\\
        \midrule
        \citet{gretz-etal-2020-large} & Quality Assessment & test & - & 1.54 & - \\ 
        & Stance Prediction & test & 0.92 & - & - \\
        \midrule
        \citet{habernal-etal-2018-argument} & Argument Reasoning Comprehension & train & - & - & 0.89\\
        \midrule
        \citet{habernal-gurevych-2016-makes} & Classify More Convincing Argument & train & 0.83 & - & - \\ 
        & Classify More Details Argument & train & 0.53 & - & - \\ 
        & Classify More Balanced Argument & train & 0.62 & - & - \\ 
        & Classify More Credible Argument & train & 0.56 & - & - \\ 
        & Classify More Clear Argument & train & 0.53 & - & - \\ 
        & Classify More On-Topic Argument & train & 0.46 & - & - \\ 
        & Classify More Provoking Argument & train & 0.57 & - & - \\ 
        & Classify More Smart Argument & train & 0.50 & - & - \\ 
        & Classify Less Attacking Argument & train & 0.75 & - & - \\ 
        & Classify Less Language-issues Argument & train & 0.66 & - & - \\ 
        & Classify Less Unclear Argument & train & 0.55 & - & - \\ 
        & Classify Less Facts Argument & train & 0.45 & - & - \\ 
        & Classify Less Reasoning Argument & train & 0.53 & - & - \\ 
        & Classify Less Relevant-reasons & train & 0.52 & - & - \\ 
        & Classify Not An Argument & train & 0.70 & - & - \\ 
        & Classify Nonsense Argument & train & 0.55 & - & - \\ 
        & Classify Off-topic Argument & train & 0.79 & - & - \\ 
        & Classify Generally weak Argument & train & 0.51 & - & - \\
        \midrule
        \citet{habernal-gurevych-2017-argumentation} & Detect Persuasive Documents & train & 0.64 & - & - \\ 
        & Extract Toulmin Components & train & - & - & 0.52\\
        \midrule
        \citet{hasan-ng-2014-taking} & Reason Identification & train & - & - & 0.56\\
        \midrule
        \citet{hautli-janisz-etal-2022-qt30} & Propositional Relations Identification & test & 0.45 & - & - \\ 
        & Illocutionary Relations Identification & test & 0.23 & - & - \\
        \midrule
        \citet{heinisch-etal-2022-overview} & Novelty Classification & train & 0.53 & - & - \\ 
        & Validity Classification & train & 0.75 & - & - \\ 
        & Relative Novelty Classification & train & 0.39 & - & - \\ 
        & Relative Validity Classification & train & 0.43 & - & - \\
        \midrule
        \citet{kuznetsov-etal-2022-revise} & Pragmatic Tagging & test & - & - & 0.24\\
        \midrule
        \citet{peldszus-2015-annotated} & Argumentative Role Determination & train & 0.79 & - & - \\ 
        & Function of Segment Determination & train & 0.36 & - & - \\ 
        & Unit Attachment Identification & train & 0.63 & - & - \\ 
        & Argumentative Text Creation & train & - & - & 0.21\\ 
        & Central Claim Extraction & train & - & - & 0.79\\
        \midrule
        \citet{persing-ng-2015-modeling} & Classifying Argument Strength & train & - & 2.05 & - \\
        \bottomrule
    \end{tabular}
    \caption{Performance of our \emph{ArgInstruct} model on all 105 CA seed tasks. The split indicates whether the task was seen during training (train) or is a completely unseen task (test). The performance is always measured on the 100 sampled instances from the test split of the respective task data. (Part 1/2)}
    \label{tab:arginstruct-eval-per-task}
\end{table*}

\begin{table*}[t]
    \centering
    \small
    \renewcommand{\arraystretch}{1}
    \setlength{\tabcolsep}{5pt}
    \begin{tabular}{llllccc}
        \toprule
        \bf Source & \bf Task & \bf Split & \bf F$_1$ $\uparrow$ & \bf MASE $\downarrow$ & \bf R-L $\uparrow$ \\
        \midrule
        \citet{poudyal-etal-2020-echr} & Argument Clause Recognition & train & 0.60 & - & - \\ 
        & Clause Relation Prediction & train & 0.58 & - & - \\ 
        & Premise Recognition & train & 0.52 & - & - \\ 
        & Conclusion Recognition & train & 0.66 & - & - \\
        \midrule
        \citet{reimers-etal-2019-classification} & Argument Similarity & train & 0.51 & - & - \\
        \midrule
        \citet{schiller-etal-2021-aspect} & Aspect Controlled Argument Generation & test & - & - & 0.11\\
        \midrule
        \citet{skitalinskaya-etal-2021-learning} & Claim Revision Improvement & train & 0.51 & - & - \\ 
        & Suboptimal Claim Detection & train & 0.50 & - & - \\ 
        & Claim Improvement Suggestions & train & 0.37 & - & - \\ 
        & Claim Optimization & train & - & - & 0.72\\
        \midrule
        \citet{stab-etal-2018-cross} & Argument Identification & train & 0.70 & - & - \\
        \midrule
        \citet{stab-gurevych-2017-parsing} & Identifying Argumentative Relations & test & 0.67 & - & - \\ 
        & Stance Recognition & test & 0.70 & - & - \\ 
        & Identifying Argument Components & test & - & - & 0.67\\ 
        & Classifying Argument Components & test & - & - & 0.67\\
        \midrule
        \citet{stahl-etal-2023-mind} & Detect Enthymemes & train & 0.50 & - & - \\ 
        & Reconstruct Enthymemes & train & - & - & 0.17\\
        \midrule
        \citet{stein-etal-2021-same} & Same Side Stance Classification & train & 0.54 & - & -\\  
        \midrule
        \citet{syed-etal-2021-generating} & Conclusion Generation & test & - & - & 0.21\\
        \midrule
        \citet{wachsmuth-etal-2017-computational} & Rate Local Acceptability & train & - & 0.95 & - \\ 
        & Rate Local Relevance & train & - & 0.88 & - \\ 
        & Rate Local Sufficiency & train & - & 0.94 & - \\ 
        & Rate Cogency & train & - & 1.00 & - \\ 
        & Rate Credibility & train & - & 0.55 & - \\ 
        & Rate Emotional Appeal & train & - & 0.83 & - \\ 
        & Rate Clarity & train & - & 0.99 & - \\ 
        & Rate Appropriateness & train & - & 0.63 & - \\ 
        & Rate Arrangement & train & - & 1.21 & - \\ 
        & Rate Effectiveness & train & - & 1.05 & - \\ 
        & Rate Global Acceptability & train & - & 1.11 & - \\ 
        & Rate Global Relevance & train & - & 0.87 & - \\ 
        & Rate Global Sufficiency & train & - & 1.02 & - \\ 
        & Rate Reasonableness & train & - & 0.96 & - \\ 
        & Rate Overall Quality & train & - & 0.88 & - \\         
        \midrule
        \citet{wachsmuth-etal-2018-argumentation} & Synthesize Argument & train & - & - & 0.20\\
        \midrule
        \citet{wachsmuth-etal-2018-retrieval} & Same Debate Opposing Counters & test & - & - & 0.46\\ 
        & Same Debate Counters & test & - & - & 0.24\\ 
        & Same Debate Opposing Argument & test & - & - & 0.30\\ 
        & Same Aebate Argument & test & - & - & 0.19\\
        \midrule
        \citet{ziegenbein-etal-2023-modeling} & Inappropriateness Detection & train & 0.72 & - & - \\ 
        & Toxic Emotions Detection & train & 0.70 & - & - \\ 
        & Missing Commitment Detection & train & 0.65 & - & - \\ 
        & Missing Intelligibility Detection & train & 0.70 & - & - \\ 
        & Other Inappropriateness Detection & train & 0.80 & - & - \\ 
        & Excessive Intensity Detection & train & 0.78 & - & - \\ 
        & Emotional Deception Detection & train & 0.78 & - & - \\ 
        & Missing Seriousness Detection & train & 0.62 & - & - \\ 
        & Missing Openness Detection & train & 0.64 & - & - \\ 
        & Unclear Meaning Detection & train & 0.71 & - & - \\ 
        & Missing Relevance Detection & train & 0.76 & - & - \\ 
        & Confusing Reasoning Detection & train & 0.72 & - & - \\ 
        & Detrimental Orthography Detection & train & 0.83 & - & - \\ 
        & Reason Unclassified Detection & train & 0.62 & - & - \\     
        \bottomrule
    \end{tabular}
    \caption{Performance of our \emph{ArgInstruct} model on all 105 CA seed tasks. The split indicates whether the task was seen during training (train) or is a completely unseen task (test). The performance is always measured on the 100 sampled instances from the test split of the respective task data. (Part 2/2)}
    \label{tab:arginstruct-eval-per-task2}
\end{table*}

\section{ArgInstruct for CA Tasks: Dual General Instruction-Finetuning}
\label{app:dual-general-isntruction-finetuning}

Table~\ref{tab:generated-existing-evaluation-full} presents the CA performance of the base models and LLMs trained with \emph{ArgInstruct} and its ablations. For completeness, we also include variants using the already instruction-fine-tuned \emph{Gemma-2-9B-General} as the base model, along with additional general instruction fine-tuning (\emph{+general}) integrated into CA-specific fine-tuning. However, this does not lead to further improvements in CA performance.

\begin{table*}[t]
	\centering
	\small
	\renewcommand{\arraystretch}{1}
	\setlength{\tabcolsep}{2pt}
	\begin{tabular}{lccclcccrccccr}
		\toprule
                                        & \multicolumn{3}{c}{\bf Fine-Tuning Data}                     && \multicolumn{4}{c}{\bf (a) Unseen CA Instances}                                             && \multicolumn{4}{c}{\bf (b) Unseen CA Tasks} \\
                                        \cmidrule{2-4}                                              \cmidrule{6-9}                                                                                 \cmidrule{11-14}
        \bf Approach                    & \bf seedCA        & \bf genCA         & \bf general       && \bf F$_1$$\uparrow$ & \bf MASE$\downarrow$ & \bf R-L $\uparrow$ & \bf Rank$\downarrow$  & & \bf F$_1$$\uparrow$ & \bf MASE$\downarrow$ & \bf R-L$\uparrow$ & \bf Rank$\downarrow$\\
        \midrule
        Gemma-2-9B                  & \emptyCircle{}    & \emptyCircle{}    & \emptyCircle{}    && .45                  & 1.6                   & .39                 & 10.0                       & & .45                  & 4.2                   & .15                 & 13.7                     \\  
         + seedCA                       & \fullCircle{}     & \emptyCircle{}    & \emptyCircle{}    && \bf \underline{.65}  & 1.1                   & \bf \underline{.50} &  \bf \underline{2.3}       & & \bf \underline{.65}  & 2.5                   & .23                 &  7.7                    \\  
         + genCA                        & \emptyCircle{}    & \fullCircle{}     & \emptyCircle{}    && .51                  & 2.7                   & .45                 &  9.7                       & & .52                  & 3.0                   & .30                 &  6.7                    \\  
         + general                      & \emptyCircle{}    & \emptyCircle{}    & \fullCircle{}     && .50                  & 2.1                   & .32                 & 11.7                       & & .51                  & 2.6                   & .29                 &  8.0                    \\  
         + seedCA, genCA                & \halfCircle{}     & \halfCircle{}     & \emptyCircle{}    && .61                  & 1.6                   & .49                 &  4.3                       & & .57                  & 2.9                   & .26                 &  7.0                    \\ 
         + genCA, general               & \emptyCircle{}    & \halfCircle{}     & \halfCircle{}     && .51                  & 1.7                   & .39                 &  9.7                       & & .50                  & 3.0                   & .17                 & 10.7                    \\  
         + seedCA, general              & \halfCircle{}     & \emptyCircle{}    & \halfCircle{}     && .59                  & \bf \underline{1.0}   & .48                 &  4.3                       & & .60                  & \underline{2.3}       & .30                 &  4.3                    \\  
         \bf + seedCA, genCA, general       & \thirdCircle{}    & \thirdCircle{}    & \thirdCircle{}    && .61                  & 1.5                   & .49                 &  4.3                       & & \bf \underline{.65}  & 2.5                   & \bf \underline{.32} & \underline{2.7}         \\  

        \addlinespace       
        Gemma-2-9B-General          & \emptyCircle{}    & \emptyCircle{}    & \fullCircleGray{}    && .48                  & 2.3                   & .34                 & 12.3                       & & .50                  & 2.1                   & .24                 & 9.3                    \\  
         + seedCA                       & \fullCircle{}     & \emptyCircle{}    & \fullCircleGray{}    && \underline{.64}      & \underline{1.2}       & \underline{.49}     &  \underline{2.7}           & & .63                  & 2.1                   & \bf \underline{.32} & 5.0                    \\    
         + genCA                        & \emptyCircle{}    & \fullCircle{}     & \fullCircleGray{}    && .49                  & 2.6                   & .44                 & 10.7                       & & .52                  & 2.6                   & .30                 & 6.7                    \\
         + general                      & \emptyCircle{}    & \emptyCircle{}    & \fullCircle{}     && .51                  & 2.1                   & .35                 & 11.3                       & & .50                  & 2.3                   & .27                 & 8.7                    \\
         + seedCA, genCA (ArgInstruct)  & \halfCircle{}     & \halfCircle{}     & \fullCircleGray{}    && .57                  & 1.3                   & \underline{.49}     &  4.7                       & & \bf \underline{.65}  & \bf \underline{1.9}   & .31                 & \bf \underline{1.7}    \\ 
         + genCA, general               & \emptyCircle{}    & \halfCircle{}     & \halfCircleGray{}     && .48                  & 1.8                   & .43                 & 10.0                       & & .50                  & 2.9                   & .31                 & 9.3                    \\  
         + seedCA, general              & \halfCircle{}     & \emptyCircle{}    & \halfCircleGray{}     && .59                  & 1.5                   & \underline{.49}     &  4.7                       & & .63                  & 2.3                   & .31                 & 5.3                    \\  
         \bf + seedCA, genCA, general       & \thirdCircle{}    & \thirdCircle{}    & \thirdCircleGray{}    && .55                  & 1.3                   & .48                 &  6.0                       & & .57                  & 2.2                   & .29                 & 5.7                    \\ 
		\bottomrule
	\end{tabular}
	\caption{Full version of Table \ref{tab:generated-existing-evaluation} containing all dataset combinations for the {\it Gemma-2-9B-General} approach. Performance on unseen CA instances and unseen CA tasks: Evaluation of Gemma-2-9B trained to follow instructions on 52k instances of CA seed tasks ({\it +seedCA}), generated CA tasks ({\it +genCA}), general tasks ({\it +general}) and combinations of these. The symbols represent the proportion of fine-tuning instances coming from each source (\texttt{\protect\emptyCircle}: 0\%, \texttt{\protect\thirdCircle}: 33\%, \texttt{\protect\halfCircle}: 50\%, \texttt{\protect\fullCircle}: 100\%). \texttt{\protect\fullCircleGray} indicates that the data was used beforehand to perform general instruction fine-tuning. Performance is evaluated on (a) unseen CA instances from the training tasks and (b) unseen CA test tasks. The best values are {\bf bold}, the best per base model \underline{underlined}.}
	\label{tab:generated-existing-evaluation-full}
\end{table*}

\end{document}